\pdfoutput=1

\documentclass[11pt]{article}

\usepackage{acl}

\usepackage{times}
\usepackage{latexsym}

\usepackage[T1]{fontenc}

\usepackage[utf8]{inputenc}

\usepackage{microtype}

\usepackage{subcaption}
\usepackage{graphicx}

\usepackage{amsfonts}
\usepackage{amsthm}
\usepackage{amsmath}
\usepackage{amssymb}
\usepackage[ruled, lined, linesnumbered, commentsnumbered, longend]{algorithm2e}

\newcommand{\calS}{\mathcal{S}}
\newcommand{\calX}{\mathcal{X}}
\newcommand{\calM}{\mathcal{M}}
\newcommand{\calD}{\mathcal{D}}
\newcommand{\calA}{\mathcal{A}}
\newcommand{\calO}{\mathcal{O}}

\newcommand{\R}{\mathbb{R}}

\newcommand{\SDP}{SentDP}

\newcommand{\mname}{\calM_{\text{TD}}}
\newcommand{\tmed}{\text{T}_{\text{MED}}}
\newcommand{\argmax}[1]{\underset{#1}{\text{arg max }}}

\newcommand{\tdappx}{\widehat{\text{TD}}}
\newcommand{\technique}{\textsc{d}eep\textsc{c}andidate}
\newcommand{\MLP}[1]{$\textbf{MLP}^{#1}$}

\SetCommentSty{mycommfont}

\newenvironment{squishlist}
{ \begin{enumerate}
    \setlength{\itemsep}{0pt}
    \setlength{\parskip}{0pt}
    \setlength{\parsep}{0pt}     }
{ \end{enumerate}    }

\newcommand{\goodreads}{\emph{Good Reads}}
\newcommand{\imdb}{\emph{IMDB}}
\newcommand{\tnews}{\emph{20 News Groups}}

\newtheorem{theorem}{Theorem}[section]

\theoremstyle{definition}
\newtheorem{definition}{Definition}[section]

\title{Sentence-level Privacy for Document Embeddings}

\author{Casey Meehan \and Khalil Mrini \and Kamalika Chaudhuri \\
        UC San Diego \\
        \texttt{\{cmeehan, kmrini, kamalika\}@eng.ucsd.edu}}

\usepackage{lipsum}

\begin{document}
\maketitle
\begin{abstract}
User language data can contain highly sensitive personal content. As such, it is imperative to offer users a strong and interpretable privacy guarantee when learning from their data. In this work, we propose \SDP: pure local differential privacy at the sentence level for a single user document. We propose a novel technique, \technique, that combines concepts from robust statistics and language modeling to produce high-dimensional, general-purpose $\epsilon$-\SDP~document embeddings. This guarantees that any single sentence in a document can be substituted with any other sentence while keeping the embedding $\epsilon$-indistinguishable. Our experiments indicate that these private document embeddings are useful for downstream tasks like sentiment analysis and topic classification and even outperform baseline methods with weaker guarantees like word-level Metric DP. 
\end{abstract}

\section{Introduction} 
\label{sec:intro}

Language models have now become ubiquitous in NLP \cite{devlin2019bert, liu2019roberta, alsentzer2019publicly}, pushing the state of the art in a variety of tasks \cite{strubell2018linguistically, liu2019multi, mrini-etal-2021-recursive}. While language models capture meaning and various linguistic properties of text \cite{jawahar2019does, yenicelik2020does}, an individual's written text can include highly sensitive information. Even if such details are not needed or used, sensitive information has been found to be vulnerable and detectable to attacks \cite{pan2020privacy, attack_word_embs, carlini_attack}. Reconstruction attacks \cite{xie2021reconstruction} have even successfully broken through private learning schemes that rely on encryption-type methods \cite{huang-etal-2020-texthide}.

As of now, there is no broad agreement on what constitutes good privacy for natural language \cite{kairouz2019advances}. \citet{huang-etal-2020-texthide} argue that different applications and models require different privacy definitions. Several emerging works propose to apply Metric Differential Privacy \cite{orig_metricdp} at the word level \cite{metricdp,  mdp_low_dim, TEM, another_metric_DP, fancy_metricdp, metricDP_gumbel} . They propose to add noise to word embeddings, such that they are indistinguishable from their nearest neighbours.

At the document level, however, the above definition has two areas for improvement. First, it may not offer the level of privacy desired. Having each word indistinguishable with similar words may not hide higher level concepts in the document, and may not be satisfactory for many users. Second, it may not be very interpretable or easy to communicate to end-users, since the privacy definition relies fundamentally on the choice of embedding model to determine which words are indistinguishable with a given word. This may not be clear and precise enough for end-users to grasp.
%

\begin{figure}
	\centering
	\includegraphics[width = 0.8\linewidth]{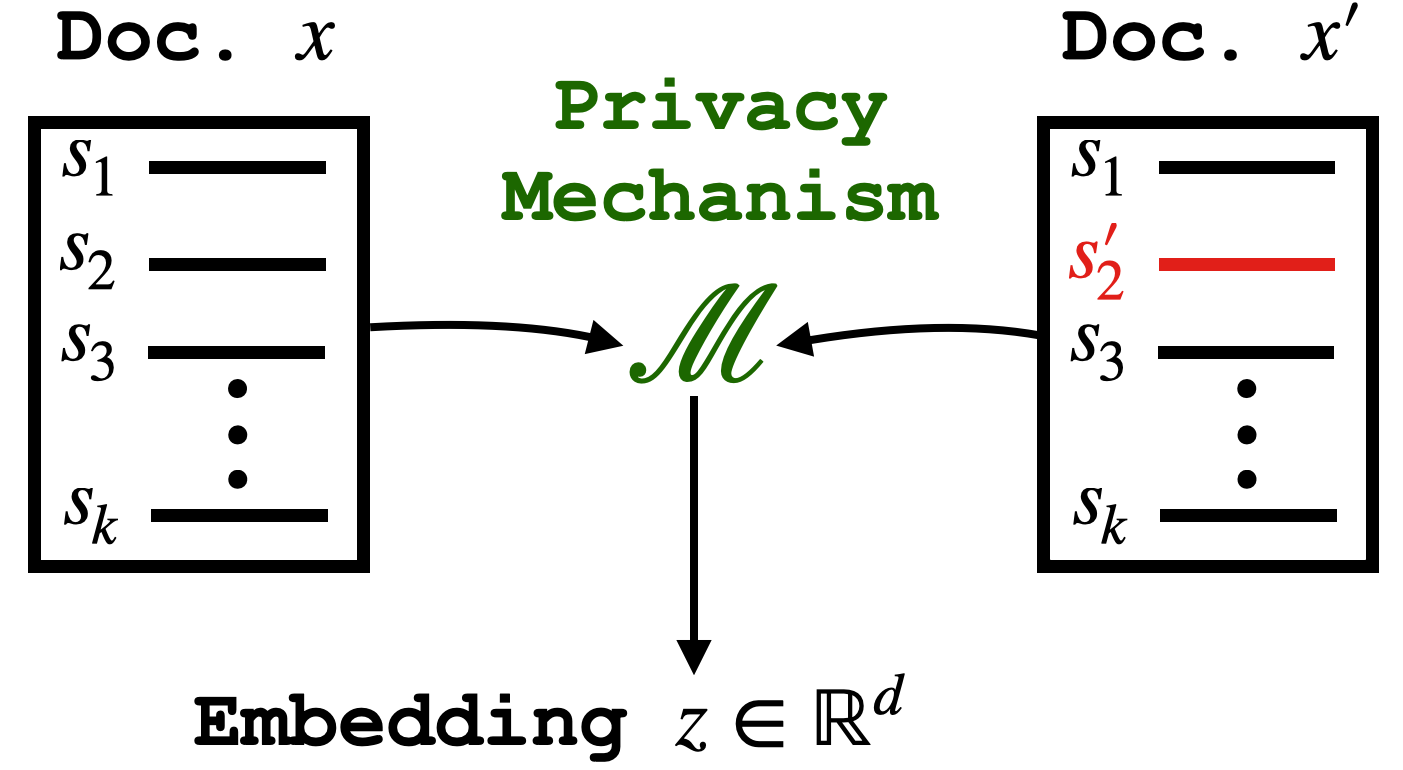}
	\label{fig:first page}
	\vspace{0cm}
	\caption{$x$ and $x'$ yield $z \in \R^d$ with similar probability.}
	\vspace{-0.5cm}
\end{figure}

\begin{figure*}
	\centering
	\vspace{-1cm}
	\includegraphics[width = \linewidth]{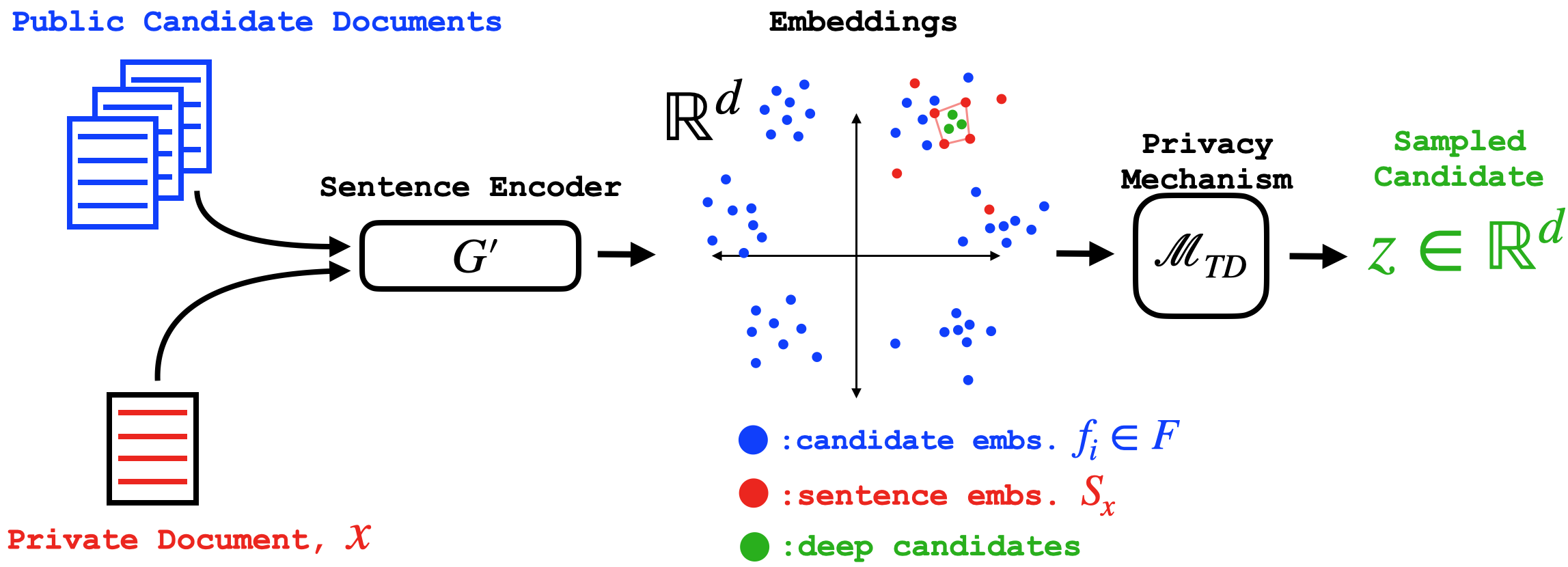}
	\label{fig:block diagram}
	\vspace{-0.65cm}
	\caption{\technique\ generates a private embedding \textcolor{green}{$z$} of document \textcolor{red}{$x$} by selecting from a set \textcolor{blue}{$F$} of public, non-private document embeddings. Sentences from \textcolor{red}{$x$} are encoded by $G'$. The privacy mechanism $\mname$, then privately samples from \textcolor{blue}{$F$}, with a preference for candidates with high Tukey Depth, `deep candidates'. $G'$ is trained beforehand to ensure that deep candidates are likely to exist and are relevant to \textcolor{red}{$x$}.}
	\vspace{-0.5cm}
\end{figure*}

In this work, we propose a new privacy definition for documents: sentence privacy. This guarantee is both strong and interpretable: any sentence in a document must be indistinguishable with \emph{any} other sentence. A document embedding is sentence-private if we can replace any single sentence in the document and have a similar probability of producing the same embedding. As such, the embedding only stores limited information unique to any given sentence. This definition is easy to communicate and strictly stronger than word-level definitions, as modifying a sentence can be changing one word.

Although this definition is strong, we are able to produce unsupervised, general embeddings of documents that are useful for downstream tasks like sentiment analysis and topic classification. To achieve this we propose a novel privacy mechanism, \technique, which privately samples a high-dimensional embedding from a preselected set of candidate embeddings derived from public, non-private data. \technique\  works by first pre-tuning a sentence encoder on public data such that semantically different document embeddings are far apart from each other. Then, we approximate each candidate's Tukey Depth within the private documents' sentence embeddings. Deeper candidates are the most likely to be sampled to represent the private document. We evaluate \technique\  on three illustrative datasets, and show that these unsupervised private embeddings are useful for both sentiment analysis and topic classification as compared to baselines. 

In summary, this work makes the following contributions to the language privacy literature:

\begin{squishlist}
	\item A new, strong, and interpretable privacy definition that offers complete indistinguishability to each sentence in a document. 
	\item A novel, unsupervised embedding technique, \technique, to generate sentence-private document embeddings. 
	\item An empirical assessment of \technique, demonstrating its advantage over baselines, delivering strong privacy and utility. 
\end{squishlist}
\section{Background and Related Work}

\paragraph{Setting.}We denote a `document' as a sequence of sentences. Let $s \in \calS$ be any finite-length sentence. Then, the space of all documents is $\calX = \calS^*$ and document $x \in \calX$ is written as $x = (s_1, s_2, \dots, s_k)$ for any non-negative integer $k$ of sentences. In this work, we focus on cohesive documents of sentences written together like reviews or emails, but our methods and guarantees apply to any sequence of sentences, such as a collection of messages written by an individual over some period of time.

Our task is to produce an embedding $z \in \R^d$ of any document $x \in \calX$ such that any single sentence $s_i \in x$ is indistinguishable with every other sentence $s_i' \in \calS \backslash s_i$. That is, if one were to replace any single sentence in the document $s_i \in x$ with \emph{any other} sentence $s_i' \in \calS \backslash s_i$, the probability of producing a given embedding $z$ is similar. To achieve this, we propose a randomized embedding function (the embedding \emph{mechanism}) $\calM : \calX \rightarrow \R^d$, that generates a private embedding $z = \calM(x)$ that is useful for downstream tasks. 

\subsection{Differential Privacy}
The above privacy notion is inspired by Differential Privacy (DP) \cite{DP}. It guarantees that --- whether an individual participates (dataset $D$) or not (dataset $D'$) --- the probability of any output only chances by a constant factor. 

\begin{definition}[Differential Privacy]
	Given any pair of datasets $D, D' \in \calD$ that differ only in the information of a single individual, we say that the mechanism $\calA : \calD \rightarrow \calO$, satisfies $\epsilon$-DP if 
	\begin{align*}
		\Pr[\calA(D) \in O] \leq e^\epsilon \Pr[\calA(D') \in O]
	\end{align*}
	for any event $O \subseteq \calO$. 
\end{definition}
Note that we take probability over the randomness of the mechanism $\calA$ only, not the data distribution. DP has several nice properties that make it easy to work with including closure under post-processing, an additive privacy budget (composition), and closure under group privacy guarantees (guarantees to a \emph{subset} of multiple participants). See \citealt{DPbook} for more details. 

%
%
%
%
%
%
The \emph{exponential mechanism} \cite{exp_mech} allows us to make a DP selection from an arbitrary output space $\calO$ based on private dataset $D$. A \emph{utility function} over input/output pairs, $u : \calD \times \calO \rightarrow \R$ determines which outputs are the best selection given dataset $D$. The log probability of choosing output $o \in \calO$ when the input is dataset $D \in \calD$ is then proportional to its utility $u(D,o)$. The \emph{sensitivity} of $u(\cdot, \cdot)$ is the worst-case change in utility over pairs of neighboring datasets $(D,D')$ that change in one entry, $\Delta u = \max_{D, D', o} | u(D,o) - u(D', o)|$. 
\begin{definition}
\label{def: exp mech} 
	The \emph{exponential mechanism} $\calA_{Exp}: \calD \rightarrow \calO$ is a randomized algorithm with output distribution
	\vspace{-0.3cm}
	\begin{align*}
	\Pr[\calA_{Exp}(D) = o] \propto \exp\big( \frac{\epsilon u(D, o)}{2 \Delta u} \big) \quad .
	\end{align*}
\end{definition}

\subsection{Related Work}
\paragraph{Natural Language Privacy.} Previous work has demonstrated that NLP models and embeddings are vulnerable to reconstruction attacks \cite{carlini_attack, attack_word_embs, pan2020privacy}. In response there have been various efforts to design privacy-preserving techniques and definitions across NLP tasks. A line of work focuses on how to make NLP model training satisfy DP \cite{DP_training, DP_training_II}. This is distinct from our work in that it satisfies central DP -- where data is first aggregated non-privately and then privacy preserving algorithms (i.e. training) are run on that data. We model this work of the \emph{local} version of DP \cite{ldp}, wherein each individual's data is made private before centralizing. Our definition guarantees privacy to a single document as opposed to a single individual. 

A line of work more comparable to our approach makes documents locally private by generating a randomized version of a document that satisfies some formal privacy definition. As with the private embedding of our work, this generates locally private \emph{representation} of a given document $x$. The overwhelming majority of these methods satisfy an instance of Metric-DP \cite{orig_metricdp} at the word level \cite{metricdp,  mdp_low_dim, TEM, another_metric_DP, fancy_metricdp, metricDP_gumbel}. As discussed in the introduction, this guarantees that a document $x$ is indistinguishable with any other document $x'$ produced by swapping a single word in $x$ with a similar word. Two words are `similar' if they are close in the word embeddings space (e.g. GloVe). This guarantee is strictly weaker than our proposed definition, \SDP, which offers indistinguishability to any two documents that differ in an entire sentence. 

\paragraph{Privacy-preserving embeddings.} There is a large body of work on non-NLP privacy-preserving embeddings, as these embeddings have been shown to be vulnerable to attacks \cite{attack_on_embeddings}. \citet{clifton} attempt to generate locally private embeddings by bounding the embedding space, and we compare with this method in our experiments. \citet{kamath_high_dim} propose a method for privately publishing the average of embeddings, but their algorithm is not suited to operate on the small number of samples (sentences) a given document offers. Finally, \citet{private_halfspaces} propose a method for privately learning halfspaces in $\R^d$, which is relevant to private Tukey Medians, but their method would restrict input examples (sentence embeddings) to a finite discrete set in $\R^d$, a restriction we cannot tolerate.

\section{Sentence-level Privacy}
We now introduce our simple, strong privacy definition, along with concepts we use to satisfy it. 
\subsection{Definition}
In this work, we adopt the \emph{local} notion of DP \cite{ldp}, wherein each individual's data is guaranteed privacy locally before being reported and centralized. Our mechanism $\calM$ receives a single document from a single individual, $x \in \calX$. We require that $\calM$ provides indistinguishability between documents $x, x'$ differing \emph{in one sentence}. 

\begin{definition}[Sentence Privacy, \SDP]
Given any pair of documents $x, x' \in \calX$ that differ only in one sentence, we say that a mechanism\\ $\calM : \calX \rightarrow \calO$ satisfies $\epsilon$-\SDP~if 
	\begin{align*}
		\Pr[\calM(x) \in O] \leq e^\epsilon \Pr[\calM(x') \in O]
	\end{align*}
	for any event $O \subseteq \calO$. 
\end{definition}

We focus on producing an embedding of the given document $x$, thus the output space is $\calO = \R^d$. For instance, consider the neighboring documents $x = (s_1, s_2, \dots, s_k)$ and $x' = (s_1, s_2', \dots, s_k)$ that differ in the second sentence, i.e. $s_2, s_2'$ can be \emph{any} pair of sentences in $\calS^2$. 
This is a strong notion of privacy in comparison to existing definitions across NLP tasks. However, we show that we can guarantee SentDP while still providing embeddings that are useful for downstream tasks like sentiment analysis and classification. In theory, a SentDP private embedding $z$ should be able to encode any information from the document that is not unique to a small subset of sentences. For instance, $z$ can reliably encode the sentiment of $x$ as long as \emph{multiple} sentences reflect the sentiment. By the group privacy property of DP, which \SDP~maintains, two documents differing in $a$ sentences are $a\epsilon$ indistinguishable. So, if more sentences reflect the sentiment, the more $\calM$ can encode this into $z$ without compromising on privacy. 

\subsection{Sentence Mean Embeddings} 

Our approach is to produce a private version of the average of general-purpose sentence embeddings. By the post-processing property of DP, this embedding can be used repeatedly in any fashion desired without degrading the privacy guarantee. Our method makes use of existing pre-trained sentence encoding models. We denote this general sentence encoder as $G : \calS \rightarrow \R^d$. We show in our experiments that the mean of sentence embeddings,  
\begin{align}
	\overline{g}(x) = \frac{1}{k} \sum_{s_i \in x} G(s_i) \ , 
	\label{eqn: doc emb}
\end{align}
maintains significant information unique to the document and is useful for downstream tasks like classification and sentiment analysis.

We call $\overline{g}(x)$ the \emph{document embedding} since it summarizes the information in document $x$. While there exist other definitions of document embeddings \cite{yang2016hierarchical, thongtan2019sentiment, bianchi2020pre}, we decide to use averaging as it is a simple and established embedding technique \cite{bojanowski2017enriching, gupta2019better, li2020sentence}.
\subsection{Tukey Depth}
Depth is a concept in robust statistics used to describe how central a point is to a distribution. We borrow the definition proposed by \citet{tukeydepth}:

\begin{definition}
\label{def: tukey} 
	Given a distribution $P$ over $\R^d$, the Tukey Depth of a point $y \in \R^d$ is 
\begin{align*}
	\text{TD}_P(y) 
	&= \inf_{w \in \R^d} P\{y' : w \cdot (y' - y) \geq 0\} \quad. 
\end{align*} 
\end{definition}
In other words, take the hyperplane orthogonal to vector $w$, $h_w$, that passes through point $y$. Let $P_1^w$ be the probability under $P$ that a point lands on one side of $h_w$ and let $P_2^w$ be the probability that a point lands on the other side, so $P_1^w + P_2^w = 1$. $y$ is considered deep if $\min(P_1^w, P_2^w)$ is close to a half for \emph{all} vectors $w$ (and thus all $h$ passing through $y$). The \emph{Tukey Median} of distribution $P$, $\tmed(P)$, is the set of all points with maximal Tukey Depth, 
\begin{align}
	\tmed(P) = \argmax{y \in \R^d} \text{TD}_P(y) \quad .
	\label{eqn:tukey median}
\end{align}
We only access the distribution $P$ through a finite sample of i.i.d. points, $Y = \{y_1, y_2, \dots, y_n\}$. The Tukey Depth w.r.t. $Y$ is given by 
\begin{align*}
	\text{TD}_Y(y) = \inf_{w \in \R^d} |\{y' \in Y : w \cdot (y' - y) \geq 0\}| \ , 
\end{align*}
and the median, $\tmed(Y)$, maximizes the depth and is at most half the size of our sample $\big \lfloor \frac{n}{2} \big  \rfloor$. 

Generally, finding a point in $\tmed(Y)$ is hard; SOTA algorithms have an exponential dependency in dimension \cite{optimal_tukey}, which is a non-starter when working with high-dimensional embeddings. However, there are efficient approximations which we will take advantage of.

\section{\technique}
\label{sec:deepcandidate}
While useful and general, the document embedding $\overline{g}(x)$ does not satisfy \SDP. We now turn to describing our privacy-preserving technique, \technique, which generates general, $\epsilon$-\SDP~document embeddings that preserve relevant information in $\overline{g}(x)$, and are useful for downstream tasks. To understand the nontrivial nature of this problem, we first analyze why the simplest, straightfoward approaches are insufficient. 




\paragraph{Motivation.}
  Preserving privacy for high dimensional objects is known to be challenging \cite{kamath_high_dim, mdp_low_dim, DP_compression} . For instance, adding Laplace noise directly to $\overline{g}(x)$, as done to satisfy some privacy definitions \cite{metricdp, orig_metricdp}, does not guarantee \SDP~for any $\epsilon$. Recall that the embedding space is all of $\R^d$. A change in one sentence can lead to an unbounded change in $\overline{g}(x)$, since we do not put any restrictions on the general encoder $G$. Thus, no matter how much noise we add to $\overline{g}(x)$ we cannot satisfy \SDP. 

A straightforward workaround might be to simply truncate embeddings such that they all lie in a limited set such as a sphere or hypercube as done in prior work \cite{clifton, abadi}. In doing so, we bound how far apart embeddings can be for any two sentences, $\|G(s_i) - G(s_i')\|_1$, thus allowing us to satisfy \SDP~by adding finite variance noise. However, such schemes offer poor utility due to the high dimensional nature of useful document embeddings (we confirm this in our experiments). We must add noise with standard deviation proportional to the dimension of the embedding, thus requiring an untenable degree of noise for complex encoders like BERT which embed into $\R^{768}$. 

Our method has three pillars: \textbf{(1)} sampling from a candidate set of public, non-private document embeddings to represent the private document, \textbf{(2)} using the Tukey median to approximate the document embedding, and \textbf{(3)} pre-training the sentence encoder, $G$, to produce relevant candidates with high Tukey depth for private document $x$. 

\subsection{Taking advantage of public data: sampling from candidates}
Instead of having our mechanism select a private embedding $z$ from the entire space of $\R^d$, we focus the mechanism to select from a set of $m$ candidate  embeddings, $F$, generated by $m$ public, non-private documents. We assume the document $x$ is drawn from some distribution $\mu$ over documents $\calX$. For example, if we know $x$ is a restaurant review, $\mu$ may be the distribution over all restaurant reviews. $F$ is then a collection of document embeddings over $m$ publicly accessible documents $x_i \sim \mu$, 
\begin{align*}
	F = \{f_i = \overline{g}(x_i) : x_1, \dots, x_m \overset{\text{iid}}{\sim} \mu\} \ , 
\end{align*}
and denote the corresponding distribution over $f_i$ as $\overline{g}(\mu)$. By selecting candidate documents that are similar in nature to the private document $x$, we inject an advantageous inductive bias into our mechanism, which is critical to satisfy strong privacy while preserving information relevant to $x$. 


\subsection{Approximating the document embedding:\\ \quad \quad \  The Tukey Median}
\label{sec:tukey}
We now propose a novel mechanism $\mname$, which approximates $\overline{g}(x)$ by sampling a candidate embedding from $F$. $\mname$ works by concentrating probability on candidates with high Tukey Depth w.r.t. the set of sentence embeddings $S_x = \{G(s_i) : s_i \in x\}$. We model sentences $s_i$ from document $x$ as i.i.d. draws from distribution $\nu_x$. Then, $S_x$ is $k$ draws from $g(\nu_x)$, the distribution of sentences from $\nu_x$ passing through $G$. Deep points are a good approximation of the mean under light assumptions. If $g(\nu_x)$ belongs to the set of halfspace-symmetric distributions (including all elliptic distributions e.g. Gaussians), we know that its mean lies in the Tukey Median \cite{tukey_props}. 

Formally, $\mname$ is an instance of the exponential mechanism (Definition \ref{def: exp mech}), and is defined by its utility function. We set the utility of a candidate document embedding $f_i \in F$ to be an approximation of its depth w.r.t. sentence embeddings $S_x$, 
\begin{align}
	u(x, f_i) = \tdappx_{S_x}(f_i) \quad. 
	\label{eqn:utility}
\end{align}
The approximation $\tdappx_{S_x}$, which we detail in the Appendix, is necessary for computational efficiency. If the utility of $f_i$ is high, we call it a `deep candidate' for sentence embeddings $S_x$.

The more candidates sampled (higher $m$), the higher the probability that at least one has high depth. Without privacy, we could report the deepest candidate, $z = \argmax{f_i \in F} \tdappx_{S_x}(f_i)$. However, when preserving privacy with $\mname$, increasing $m$ has diminishing returns. To see this, fix a set of sentence embeddings $S_x$ for document $x$ and the i.i.d. distribution over candidate embeddings $f_i \sim \overline{g}(\mu)$. This induces a multinomial distribution over depth,  
\vspace{-0.6cm}
\begin{align*}
	u_j(x) = \Pr[u(x, f_i) = j], \ \ \sum_{j = 0}^{\lfloor \frac{k}{2} \rfloor} u_j(x) = 1 \ ,
\end{align*}
\vspace{-0.5cm}

\noindent where randomness is taken over draws of $f_i$. 

For candidate set $F$ and sentence embeddings $S_x$, the probability of $\mname$'s selected candidate, $z$, having (approximated) depth $j^*$ is given by 
\begin{align}
	\Pr[u(x, z) = j^*] = \frac{a_{j^*}(x)e^{\epsilon j^* / 2}}{\sum_{j=0}^{\lfloor \frac{k}{2} \rfloor} a_j(x) e^{\epsilon j / 2}}
	\label{eqn:prob deep}
\end{align}
where $a_j(x)$ is the fraction of candidates in $F$ with depth $j$ w.r.t. the sentence embeddings of document $x$, $S_x$. For $m$ sufficiently large, $a_j(x)$ concentrates around $u_j(x)$, so further increasing $m$ 
does not increase the probability of $\mname$ \emph{sampling} a deep candidate. 

\begin{table}[h!]
  \begin{center}
  \vspace{-0.25cm}
    \caption{Conditions for deep candidates}
    \label{tab:mech example}
    \begin{tabular}{l|c|r} 
      $\epsilon$ & $b$ & $j^*$ \\
      \hline
      3 & 55 & 5\\
      6 & 25 & 3\\
      10 & 5 & 2\\
      23 & 1 & 1\\
    \end{tabular}
  \end{center}
  \vspace{-0.5cm}
\end{table}

For numerical intuition, suppose $m = 5000$ (as in our experiments), $\geq b$ candidates have depth $\geq j^*$, and all other candidates have depth 0, $\mname$ will sample one of these deep candidates w.p. $\geq 0.95$ under the settings in Table \ref{tab:mech example}. 

\begin{figure}
	\centering
	\includegraphics[width = \columnwidth]{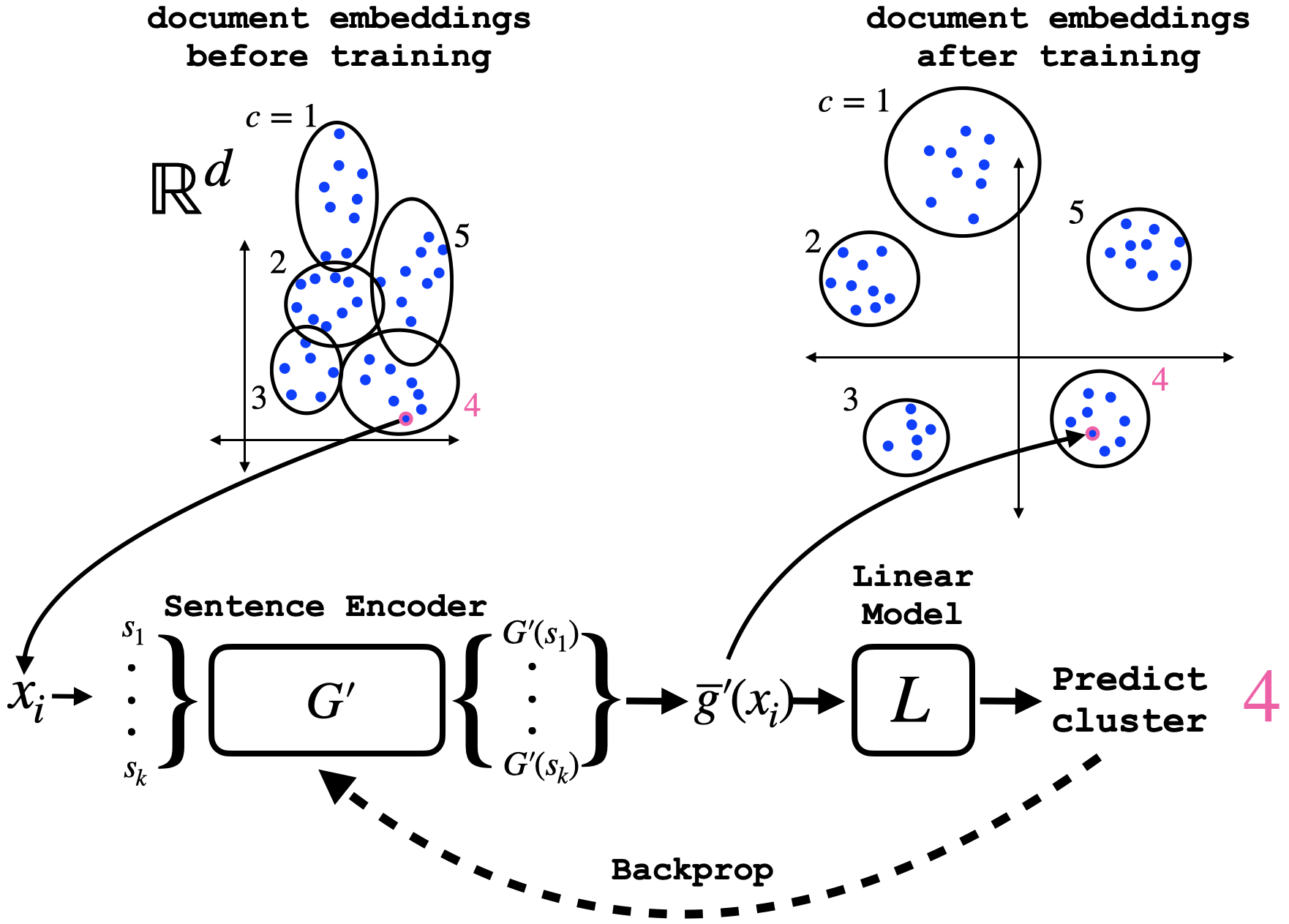} 
	\vspace{-0.65cm}
	\caption{$G'$ is trained to encourage similar documents to embed close together and different documents to embed far apart. We first compute embeddings of all (public, non-private) training set documents $T$ with pretrained encoder $G$, $T_G = \{t_i = \overline{g}(x_i) : x_i \in T\}$ (blue dots). We run $k$-means to define $n_c$ clusters, and label each training document embedding $t_i \in T_G$ with its cluster $c$. We then train $H$ to recode sentences to $S_x'$ such that their mean $\overline{g}'(x)$ can be used by a linear model $L$ to predict cluster $c$. Our training objective is the cross-entropy loss of the linear model $L$ in predicting $c$.}
	\label{fig:training diagram}
	\vspace{-0.4cm}
\end{figure}

For low $\epsilon < 10$ (high privacy), about 1\% of candidates need to have high depth $(\geq 3)$ in order to be reliably sampled. Note that this is only possible for documents with $\geq 6$ sentences. For higher $\epsilon \geq 10$, $\mname$ will reliably sample low depth candidates even if there are only a few. 
 
From these remarks we draw two insights on how \technique\ can achieve high utility.\\
\textbf{(1)} \emph{More sentences} A higher $k$ enables greater depth, and thus a higher probability of sampling deep candidates with privacy. We explore this effect in our experiments. \\
\textbf{(2)} \emph{ Tuned encoder} By tuning the sentence encoder $G$ for a given domain, we can modify the distribution over document embeddings $\overline{g}(\mu)$ and sentence embeddings $g(\nu_x)$ to encourage deep candidates (high probability $u_j$ for deep $j$) that are relevant to document $x$.

\begin{figure*}[ht]
\vspace{-0.75cm}
\centering 
   \begin{subfigure}[b]{0.30\linewidth}
       \centering
       \includegraphics[width=0.95\linewidth]{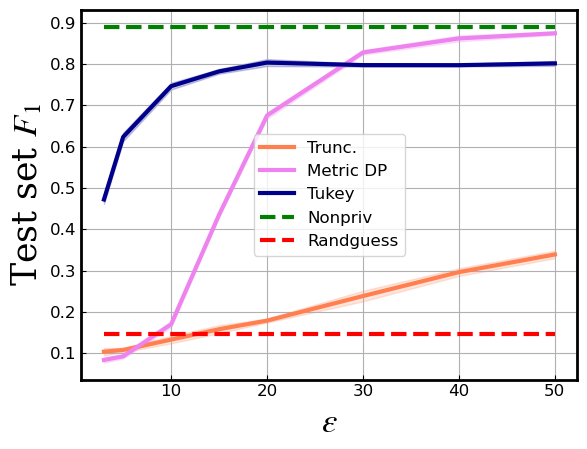}
      \vspace{-0.15cm}
       \caption{\textit{20 News}: Sweep $\epsilon$}
       \label{fig:eps:tnews}
    \end{subfigure}
    \begin{subfigure}[b]{0.30\linewidth}
       \centering
       \includegraphics[width=0.95\linewidth]{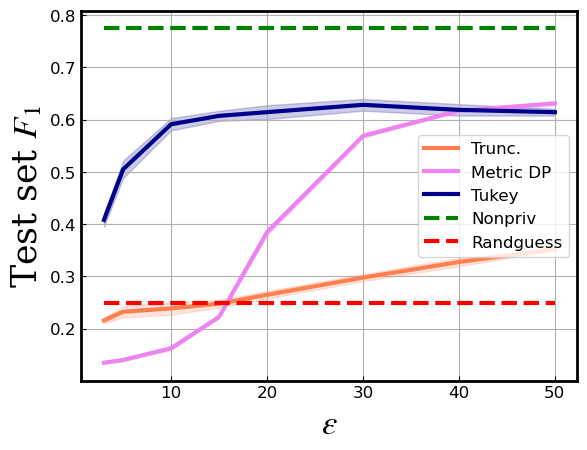}
      \vspace{-0.15cm}
       \caption{\textit{GoodReads}: Sweep $\epsilon$}
       \label{fig:eps:gr}
    \end{subfigure}
    \begin{subfigure}[b]{0.30\linewidth}
       \centering
       \includegraphics[width=0.95\linewidth]{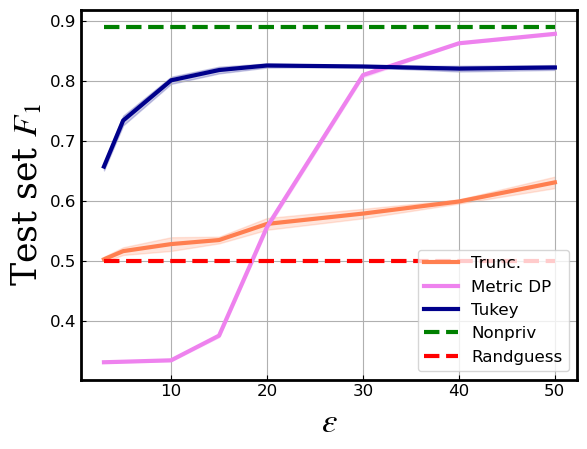}
      \vspace{-0.15cm}
       \caption{\textit{IMDB}: Sweep $\epsilon$}
       \label{fig:eps:imdb}
    \end{subfigure}
    \hfill
    \begin{subfigure}[b]{0.30\linewidth}
       \centering
       \includegraphics[width=0.95\linewidth]{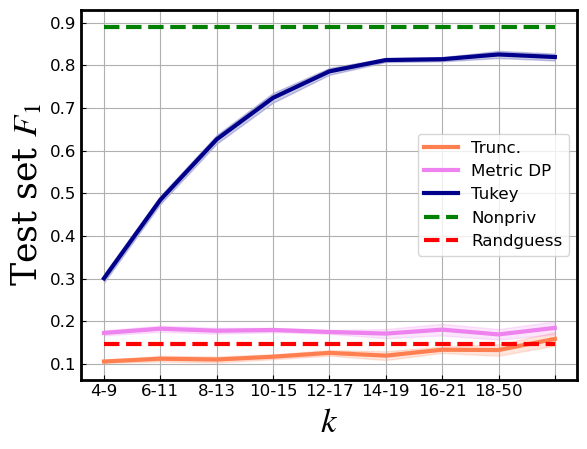}
      \vspace{-0.15cm}
       \caption{\textit{20 News}: Sweep $k$}
       \label{fig:k:tnews}
    \end{subfigure}
    \begin{subfigure}[b]{0.30\linewidth}
       \centering
       \includegraphics[width=0.95\linewidth]{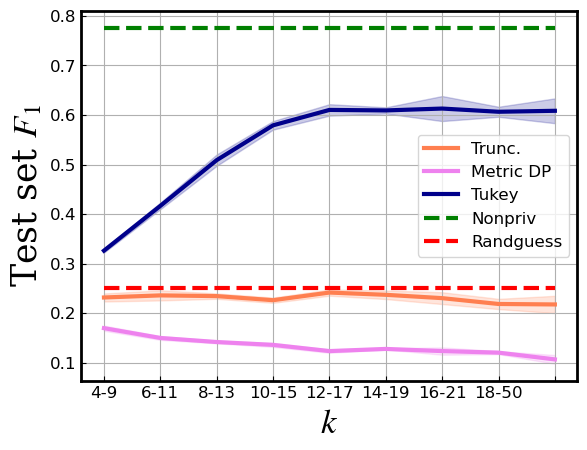}
      \vspace{-0.15cm}
       \caption{\textit{GoodReads}: Sweep $k$}
       \label{fig:k:gr}
    \end{subfigure}
    \begin{subfigure}[b]{0.30\linewidth}
       \centering
       \includegraphics[width=0.95\linewidth]{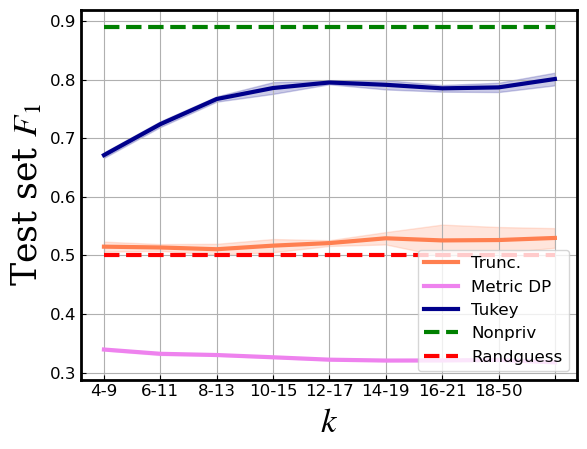}
      \vspace{-0.15cm}
       \caption{\textit{IMDB}: Sweep $k$}
       \label{fig:k:imdb}
    \end{subfigure}
    \caption{Comparison of our mechanism with two baselines: truncation \cite{clifton} and word-level Metric DP \cite{metricdp} for both sentiment analysis (\emph{IMDB}) and topic classification (\emph{GoodReads}, \emph{20News}) on private, unsupervised embeddings. All plots show test-set macro $F_1$ scores. The top row shows performance vs. privacy parameter $\epsilon$ (lower is better privacy). The bottom row shows performance vs. number of sentences $k$ with $\epsilon = 10$. \technique\ outperforms both baselines across datasets and tasks. Note that at a given $\epsilon$, word-level Metric-DP is a significantly weaker privacy guarantee.}
\end{figure*}

\subsection{Taking advantage of structure: cluster-preserving embeddings}

So far, we have identified that deep candidates from $F$ can approximate $\overline{g}(x)$. To produce a good approximation, we need to ensure that 1) there reliably exist deep candidates for any given set of sentence embeddings $S_x$, and 2) that these deep candidates are good representatives of document $x$. The general sentence encoder $G$ used may not satisfy this `out of the box'. If the distribution on document embeddings $\overline{g}(\mu)$ is very scattered around the instance space $\R^{768}$, it can be exceedingly unlikely to have a deep candidate $f_i$ among sentence embeddings $S_x$. On the other hand, if distribution $\overline{g}(\mu)$ is tightly concentrated in one region (e.g. `before training' in Figure \ref{fig:training diagram}), then we may reliably have many deep candidates, but several will be poor representatives of the document embedding $\overline{g}(x)$. 

To prevent this, we propose an unsupervised, efficient, and intuitive modification to the (pretrained) sentence encoder $G$. We freeze the weights of $G$ and add additional perceptron layers mapping into the same embeddings space $H:\R^d \rightarrow \R^d$, producing the extended encoder $G' = H \circ G$. Broadly, we train $H$ to place similar document embeddings close together, and different embeddings far part.  To do so, we leverage the assumption that a given domain's distribution over document embeddings $\overline{g}(\mu)$ can be parameterized by $n_c$ clusters, visualized as the black circles in Figure \ref{fig:training diagram}. $H$'s aim is to recode sentence embeddings such that document embedding clusters are preserved, but spaced apart from each other. By preserving clusters, we are more likely to have deep candidates (increased probability $u_j$ for high depth $j$). By spacing clusters apart, these deep candidates are more likely to come from the same or a nearby cluster as document $x$, and thus be good representatives. Note that $H$ is domain-specific: we train separate $H$ encoders for each dataset.

\subsection{Sampling Algorithm}
The final component of \technique\ is computing the approximate depth of a candidate for use as utility in the exponential mechanism as in Eq. \eqref{eqn:utility}. We use a version of the approximation algorithm proposed in \citealt{median_hyp}. Intuitively, our algorithm computes the one-dimensional depth of each $f_i$ among $x$'s sentence embeddings $S_x$ on each of $p$ random projections. The approximate depth of $f_i$ is then its lowest depth across the $p$ projections. We are guaranteed that $\tdappx_{S_x}(f_i) \geq \text{TD}_{S_x}(f_i)$. Due to space constraints, we leave the detailed description of the algorithm for the Appendix.
\begin{theorem}
\label{thm:mainthm}
	$\mname$ satisfies $\epsilon$-Sentence Privacy
\end{theorem}
Proof follows from the fact that $\tdappx_{S_x}(f_i)$ has bounded sensitivity (changing one sentence can only change depth of $f_i$ by one). We expand on this, too, in the Appendix.

\section{Experiments}
\label{sec:experiments}
\vspace{-.2cm}
\subsection{Datasets}
\vspace{-.1cm}
\label{sec: datasets} 
We produce private, general embeddings of documents from three English-language datasets: 

\textbf{\goodreads} \cite{goodreads} 60k book reviews from four categories: fantasy, history, romance, and childrens literature.  Train-48k | Val-8k | Test-4k 

\textbf{\tnews} \cite{20newsgroup} 11239 correspondences from 20 different affinity groups. Due to similarity between several groups (e.g. \texttt{comp.os.ms-windows.misc} and \texttt{comp.sys.ibm.pc.hardware}), the dataset is partitioned into nine categories. Train-6743k | Val-2247k | Test-2249k

\textbf{\imdb} \cite{imdb} 29k movie reviews from the IMDB database, each labeled as a positive or negative review. Train-23k | Val-2k | Test-4k 

To evaluate utility of these unsupervised, private embeddings, we check if they are predictive of document properties. For the \goodreads\ and \tnews\ datasets, we evaluate how useful the embeddings are for topic classification. For \imdb\ we evaluate how useful the embeddings are for sentiment analysis (positive or negative review). Our metric for performance is test-set macro $F_1$ score. 

\subsection{Training Details \& Setup}
For the general encoder, $G:\calS \rightarrow \R^{768}$, we use SBERT \cite{sbert}, a version of BERT fine-tuned for sentence encoding. Sentence embeddings are generated by mean-pooling output tokens. In all tasks, we freeze the weights of SBERT. The cluster-preserving recoder, $H$, as well as every classifier is implemented as an instance of a 4-layer MLP taking $768$-dimension inputs and only differing on output dimension. We denote an instance of this MLP with output dimension $o$ as \MLP{o}. We run 5 trials of each experiment with randomness taken over the privacy mechanisms, and plot the mean along with a $\pm$ 1 standard deviation envelope. 

\paragraph{\technique:} The candidate set $F$ consists of 5k document embeddings from the training set, each containing at least 8 sentences. To train $G'$, we find $n_c = 50$ clusters with $k$-means. We train a classifier $C_{\text{dc}} = $ \MLP{r} on document embeddings $g'(x)$ to predict class, where $r$ is the number of classes (topics or sentiments). 

\subsection{Baselines}
We compare the performance of \technique\ with 4 baselines: \textbf{Non-private}, \textbf{Truncation}, \textbf{Word-level Metric-DP}, and \textbf{Random Guesser}. 

\textbf{Non-private:} This demonstrates the usefulness of non-private sentence-mean document embeddings $\overline{g}(x)$. We generate $\overline{g}(x)$ for every document using SBERT, and then train a classifier $C_{\text{nonpriv}} = $ \MLP{r} to predict $x$'s label from $\overline{g}(x)$. 

\textbf{Truncation:} We adopt the method from \citealt{clifton} to truncate (clip) sentence embeddings within a box in $\R^{768}$, thereby bounding sensitivity as described at the beginning of Section \ref{sec:deepcandidate}. Laplace noise is then added to each dimension. Documents with more sentences have proportionally less noise added due to the averaging operation reducing sensitivity. 


\textbf{Word Metric-DP (MDP):} The method from \citealt{metricdp} satisfies $\epsilon$-word-level metric DP by randomizing words. We implement MDP to produce a randomized document $x'$, compute $\overline{g}(x')$ with SBERT, and predict class using $C_{\text{nonpriv}}$. 

\textbf{Random Guess:} To set a bottom-line, we show the theoretical performance of a random guesser only knowing the distribution of labels. 

\subsection{Results \& Discussion} 
\textbf{How does performance change with privacy parameter $\epsilon$?}\\ 
This is addressed in Figures \ref{fig:eps:tnews} to \ref{fig:eps:imdb}. Here, we observe how the test set macro $F_1$ score changes with privacy parameter $\epsilon$ (a lower $\epsilon$ offers stronger privacy). Generally speaking, for local differential privacy, $\epsilon < 10$ is taken to be a strong privacy regime, $10 \leq \epsilon < 20$ is moderate privacy, and $\epsilon \geq 25$ is weak privacy. The \textbf{truncation} baseline mechanism does increase accuracy with increasing $\epsilon$, but never performs much better than the random guesser. This is to be expected with high dimension embeddings, since the standard deviation of noise added increases linearly with dimension. 

The word-level \textbf{MDP} mechanism performs significantly better than \textbf{truncation}, achieving relatively good performance for $\epsilon \geq 30$. There are two significant caveats, however. First, is the privacy definition: as discussed in the Introduction, for the same $\epsilon$, word-level MDP is strictly weaker than \SDP. 
The second caveat is the level of $\epsilon$ at which privacy is achieved. Despite a weaker privacy definition, the MDP mechanism does not achieve competitive performance until the weak-privacy regime of $\epsilon$. We suspect this is due to two reasons. First, is the fact that the MDP mechanism does not take advantage of contextual information in each sentence as our technique does; randomizing each word independently does not use higher level linguistic information. Second, is the fact that the MDP mechanism does not use domain-specific knowledge as our mechanism does with use of relevant candidates and domain specific sentence encodings. 

In comparison, \technique\ offers strong utility across tasks and datasets for relatively low values of $\epsilon$, even into the strong privacy regime. Beyond $\epsilon = 25$, the performance of \technique\ tends to max out, approximately 10-15\% below the non-private approach. This is due to the fact that \technique\ offers a noisy version of an \emph{approximation} of the document embedding $\overline{g}(x)$ -- it cannot perform any better than deterministically selecting the deepest candidate, and even this candidate may be a poor representative of $x$. We consider this room for improvement, since there are potentially many other ways to tune $G'$ and select the candidate pool $F$ such that deep candidates are nearly always good representatives of a given document $x$. 

\noindent\textbf{How does performance change with the number of sentences $k$?}\\
This is addressed in Figures \ref{fig:k:tnews} to \ref{fig:k:imdb}. We limit the test set to those documents with $k$ in the listed range on the x-axis. We set $\epsilon = 10$, the limit of the strong privacy regime. Neither baseline offers performance above that of the random guesser at this value of $\epsilon$.  \technique\ produces precisely the performance we expect to see: documents with more sentences result in sampling higher quality candidates, confirming the insights of Section \ref{sec:tukey}. Across datasets and tasks, documents with more than 10-15 sentences tend to have high quality embeddings. 

\section{Conclusions and Future Work}
\vspace{-0.5em}
We introduce a strong and interpretable local privacy guarantee for documents, \SDP, along with \technique, a technique that combines principles from NLP and robust statistics to generate general $\epsilon$-\SDP\ embeddings. Our experiments confirm that such methods can outperform existing approaches even with with more relaxed privacy guarantees. Previous methods have argued that it is ``virtually impossible'' to satisfy pure local DP \cite{metricdp, mdp_low_dim} at the word level while capturing linguistic semantics. Our work appears to refute this notion at least at the document level. 

To follow up, we plan to explore other approaches (apart from $k$-means) of capturing the structure of the embedding distribution $\overline{g}(\mu)$ to encourage better candidate selection. We also plan to experiment with decoding private embeddings back to documents by using novel candidates produced by a generative model trained on $F$. 


\section*{Acknowledgements} 
KC and CM would like to thank ONR under N00014-20-1-2334. KM gratefully acknowledges funding from an Amazon Research Award and Adobe Unrestricted Research Gifts. We would would also like to thank our reviewers for their insightful feedback.
%



\bibliography{anthology,custom}
\bibliographystyle{acl_natbib}

\clearpage 

\appendix

\section{Appendix} 
\label{sec:appendix} 

\subsection{Privacy Mechanism}
We now describe in detail our instance of the exponential mechanism $\mname$. Recall from Definition \ref{def: exp mech} that the exponential mechanism samples candidate $f_i \in F$ with probability
\begin{align*}
	\Pr[\calM(x) = f_i] \propto \exp\big( \frac{\epsilon u(x, f_i)}{2 \Delta u} \big) \ .
\end{align*}
Thus, $\mname$ is fully defined by its utility function, which, as listed in Equation \eqref{eqn:utility}, is approximate Tukey Depth, 
\begin{align*}
u(x, f_i) = \tdappx_{S_x}(f_i) \quad.
\end{align*}
We now describe our approximation algorithm of Tukey Depth $\tdappx_{S_x}(f_i)$, which is an adaptation of the general median hypothesis algorithm proposed by \citet{median_hyp}. 

\begin{algorithm}
    \SetKwFunction{isOddNumber}{isOddNumber}
    \SetKwInOut{KwIn}{Input}
    \SetKwInOut{KwOut}{Output}

    \KwIn{$m$ candidates $F$, \\sentence embs. $S_x = (s_1, \dots, s_k)$,\\ number of projections $p$}
    \KwOut{probability of sampling each candidate $P_F = [P_{f_1}, \dots, P_{f_m}]$}
    
    $v_1, \dots, v_p \gets $ random vecs. on unit sphere 
    
    \tcp{Project all embeddings}
  
    \For{$i \in [k]$}{
    \For{$j \in [p]$}{
    $s_i^j \gets s_i^\intercal v_j$
    }
    }
    
    \For{$i \in [m]$}{
    \For{$j \in [p]$}{
    $f_i^j \gets f_i^\intercal v_j$
    
    \tcc{Compute depth of $f_i$ on projection $v_j$}
    
    $h_j(x,f_i) \gets \#\{s_l^j : s_l^j \geq  f_i^j, l \in [k]\}$
    
    $u_j(x,f_i) \gets -\big| h_j(x,f_i) - \frac{k}{2} \big|$ 
    }
    $u(x,f_i) \gets \max_{j \in [p]} u_j(x,f_i)$
    $\hat{P}_{f_i} \gets \exp(\epsilon u(x,f_i) / 2)$
    }
    
    $\Psi \gets \sum_{i=1}^{m} \hat{P}_{f_i}$
    
    \For{$i \in [m]$}{
    $P_{f_i} \gets \frac{1}{\Psi} \hat{P}_{f_i}$
    }
    
    \KwRet{$P_F$}
    \caption{$\mname$ compute probabilities}
    \label{alg:main alg}
\end{algorithm}

Note that we can precompute the projections on line 10. The runtime is $O(mkp)$: for each of $m$ candidates and on each of $p$ projections, we need to compute the scalar difference with $k$ sentence embeddings. Sampling from the multinomial distribution defined by $P_F$ then takes $O(m)$ time. 

Additionally note from lines 13 and 15 that utility has a maximum of 0 and a minimum of $-\frac{k}{2}$, which is a semantic change from the main paper where maximum utility is $\frac{k}{2}$ and minimum is 0. 

\subsection{Proof of Privacy}

\textbf{Theorem \ref{thm:mainthm}} \emph{
	$\mname$ satisfies $\epsilon$-Sentence Privacy
}
\begin{proof}

	It is sufficient to show that the sensitivity, 
	\begin{align*}
		\Delta u = \max_{x, x', f_i} | u(x,f_i) - u(x', f_i)| \leq 1 \quad . 
	\end{align*} 
	Let us expand the above expression using the terms in Algorithm \ref{alg:main alg}. 
	\begin{align*}
		\Delta u &= \max_{x, x', f_i} | \max_{j \in [p]} u_j(x,f_i)  - \max_{j' \in [p]} u_{j'}(x',f_i)| \\ 
		&= \max_{x, x', f_i} | \min_{j \in [p]} \big| h_j(x,f_i) - \frac{k}{2} \big|  \\
		&- \min_{j' \in [p]} \big| h_{j'}(x',f_i) - \frac{k}{2} \big|| \\
		&\leq \max_{ f_i} | \min_{j \in [p]} \big| h_j(x,f_i) - \frac{k}{2} \big|  \\
		&- \big( \min_{j' \in [p]} \big| h_{j'}(x,f_i) - \frac{k}{2} \big|-1\big) | \\
		&\leq 1
	\end{align*}
	The last step follows from the fact that $|h_j(x, f_i) - h_j(x', f_i)| \leq 1$ for all $j \in [p]$. In other words, by modifying a single sentence embedding, we can only change the number of embeddings greater than $f_i^j$ on projection $j$ by 1. So, the distance of $h_j(x, f_i)$ from $\frac{k}{2}$ can only change by 1 on each projection. In the `worst case', the distance $\big| h_j(x,f_i) - \frac{k}{2} \big|$ reduces by 1 on every projection $v_j$. Even then, the minimum distance from $\frac{k}{2}$ across projections (the worst case depth) can only change by 1, giving us a sensitivity of 1. 
\end{proof}


\subsection{Experimental Details}

Here, we provide an extended, detailed version of section \ref{sec:experiments}. 

For the general encoder, $G:\calS \rightarrow \R^{768}$, we use SBERT \cite{sbert}, a version of BERT fine-tuned for sentence encoding. Sentence embeddings are generated by mean-pooling output tokens. In all tasks, we freeze the weights of SBERT. The cluster-preserving recoder, $H$, as well as every classifier is implemented as an instance of a 4-layer MLP taking $768$-dimension inputs and only differing on output dimension. We denote an instance of this MLP with output dimension $o$ as \MLP{o}. We run 5 trials of each experiment with randomness taken over the privacy mechanisms, and plot the mean along with a $\pm$ 1 standard deviation envelope. 

\paragraph{Non-private:} For our non-private baseline, we demonstrate the usefulness of sentence-mean document embeddings. First, we generate the document embeddings $\overline{g}(x_i)$ for each training, validation, and test set document using SBERT, $G$. We then train a classifier $C_{\text{nonpriv}} = $ \MLP{r} to predict each document's topic or sentiment, where $r$ is the number of classes. The number of training epochs is determined with the validation set. 

\paragraph{\technique :} We first collect the candidate set $F$ by sampling 5k document embeddings from the subset of the training set containing at least 8 sentences. We run $k$-means with $n_c = 50$ cluster centers, and label each training set document embedding $t_i \in T_G$ with its cluster. The sentence recoder, $H = $ \MLP{768} is trained on the training set along with the linear model $L$ with the Adam optimizer and cross-entropy loss. For a given document $x$, its  sentence embeddings $S_x$ are passed through $H$, averaged together, and then passed to $L$ to predict $x$'s cluster. $L$'s loss is then back-propagated through $H$. A classifier $C_{\text{dc}} = $ \MLP{r} is trained in parallel using a separate instance of the Adam optimizer to predict class from the recoded embeddings, where $r$ is the number of classes (topics or sentiments). The number of training epochs is determined using the validation set. At test time, (generating private embeddings using $\mname$), the optimal number of projections $p$ is empirically chosen for each $\epsilon$ using the validation set. 

\paragraph{Truncation:} The truncation baseline \cite{clifton} requires first constraining the embedding instance space. We do so by computing the 75\% median interval on each of the 768 dimensions of training document embeddings $T_G$. Sentence embeddings are truncated at each dimension to lie in this box. In order to account for this distribution shift, a new classifier $C_{\text{trunc}} = $ \MLP{r} is trained on truncated mean embeddings to predict class. The number of epochs is determined with the validation set. At test time, a document's sentence embeddings $S_x$ are truncated and averaged. We then add Laplace noise to each dimension with scale factor $\frac{768 w}{k \epsilon}$, where $w$ is the width of the box on that dimension (\emph{sensitivity} in DP terms). Note that the standard deviation of noise added is inversely proportional to the number of sentences in the document, due to the averaging operation reducing sensitivity. 

\paragraph{Word Metric-DP:} Our next baseline satisfies $\epsilon$-word-level metric DP and is adopted from \cite{metricdp}. The corresponding mechanism $\text{MDP}: \calX \rightarrow \calX$ takes as input a document $x$ and returns a private version, $x'$, by randomizing each word individually. For comparison, we generate document embeddings by first randomizing the document $x' = \text{MDP}(x)$ as prescribed by \cite{metricdp}, and then computing its document embedding $\overline{g}(x')$ using SBERT. At test time, we classify the word-private document embedding using $C_{\text{nonpriv}}$. 

\paragraph{Random Guess:} To set a bottom-line, we show the theoretical performance of a random guesser. The guesser chooses class $i$ with probability $q_i$ equal to the fraction of $i$ labels in the training set. The performance is then given by $\sum_{i = 1}^{r} q_i^2$.

\subsection{Reproducability Details}
We plan to publish a repo of code used to generate the exact figures in this paper (random seeds have been set) with the final version. Since we do not train the BERT base model $G$, our algorithms and training require relatively little computational resouces. Our system includes a single Nvidia GeForce RTX 2080 GPU and a single Intel i9 core. All of our models complete an epoch training on all datasets in less than one minute. We never do more than 20 epochs of training. All of our classifier models train (including linear model) have less than 11 million parameters. The relatively low amount of parameters is due to the fact that we freeze the underlying language model. The primary hyperparameter tuned is the number of projections $p$. We take the argmax value on the validation set between 10 and 100 projections. We repeat this for each value of $\epsilon$. 

\paragraph{Dataset preprocessing:} For all datasets, we limit ourselves to documents with at least 2 sentences. 

\imdb: This dataset has pre-defined train/test splits. We use the entire training set and form the test set by randomly sampling 4,000 from the test set provided. We do this for efficiency in computing the Metric-DP baseline, which is the slowest of all algorithms performed. Since the Metric-DP baseline randomizes first, we cannot precompute the sentence embeddings $G(s_i)$ -- we need to compute the sentence embeddings every single time we randomize. Since we randomize for each sentence of each document at each $\epsilon$ and each $k$ over 5 trials -- this takes a considerable amount of time. 

\goodreads: This dataset as provided is quite large. We randomly sample 15000 documents from each of 4 classes, and split them into 12K training examples, 2K validation examples, and 1K test examples per class. 

\tnews: We preprocess this dataset to remove all header information, which may more directly tell information about document class, and only provide the model with the sentences from the main body. We use the entire dataset, and form the Train/Val/Test splits by random sampling.

\end{document}